\documentclass{article}

\usepackage[dblblindworkshop, final]{neurips_2025}

\usepackage[utf8]{inputenc} 
\usepackage[T1]{fontenc}    
\usepackage{hyperref}       
\usepackage{url}            
\usepackage{booktabs}       
\usepackage{amsfonts}       
\usepackage{nicefrac}       
\usepackage{microtype}      
\usepackage{xcolor}         
\usepackage{graphicx}       
\usepackage{amsmath}        
\usepackage{amssymb}        
\usepackage{tikz}
\usetikzlibrary{shapes.geometric, arrows.meta, positioning, calc}

\workshoptitle{MATH-AI: The 5th Workshop on Mathematical Reasoning and AI}

\title{Automated Discovery of Conservation Laws via Hybrid Neural ODE-Transformers}

\author{%
  Vivan Doshi \\
  Independent Researcher \\
  San Jose, CA 95148 \\
  \texttt{vivandoshi24@gmail.com}
}

\begin{document}

\maketitle

\begin{abstract}
The discovery of conservation laws is a cornerstone of scientific progress. However, identifying these invariants from observational data remains a significant challenge. We propose a hybrid framework to automate the discovery of conserved quantities from noisy trajectory data. Our approach integrates three components: (1) a Neural Ordinary Differential Equation (Neural ODE) that learns a continuous model of the system's dynamics, (2) a Transformer that generates symbolic candidate invariants conditioned on the learned vector field, and (3) a symbolic-numeric verifier that provides a strong numerical certificate for the validity of these candidates. We test our framework on canonical physical systems and show that it significantly outperforms baselines that operate directly on trajectory data. This work demonstrates the robustness of a decoupled learn-then-search approach for discovering mathematical principles from imperfect data.
\end{abstract}

\section{Introduction}
As science enters a "third paradigm" of data-driven discovery \cite{kramer2023automated}, a central challenge is distilling fundamental principles from complex datasets. Among the most profound are conservation laws—invariants reflecting a system's underlying symmetries. Automating this process is profoundly difficult, as real-world data is often noisy, sparse, and irregularly sampled, obscuring the underlying physical laws.

Existing methods often fall into two camps. On one hand, models like Neural ODEs \cite{chen2019neural} learn continuous-time dynamics with high fidelity but yield opaque representations. On the other hand, symbolic regression techniques like SINDy \cite{brunton2016sindy} and AI Feynman \cite{udrescu2020aifeynman} search for simple expressions but can be brittle to noise.

We propose a hybrid framework that synergizes these approaches. Our novelty lies in a specific three-stage pipeline: we first learn a continuous vector field, then condition a symbolic search on this learned model, and finally use a rigorous numerical verifier to filter candidates. Our core contributions are:
\begin{itemize}
    \vspace{-\topsep}
    \item A decoupled architecture that first learns a system's vector field with a Neural ODE, then uses a Transformer to generate symbolic candidates based on this learned model.
    \item A symbolic-numeric verification module that acts as a strong filter, ensuring candidates are true invariants of the learned dynamics, not artifacts of data noise.
    \item An empirical demonstration that this pipeline significantly outperforms end-to-end baselines in discovering known conservation laws from noisy data.
    \vspace{-\topsep}
\end{itemize}

\section{Related Work}
Our approach synthesizes ideas from several research domains.

\textbf{Dynamics Learning and Symbolic Regression:} Neural ODEs \cite{chen2019neural} provide a powerful framework for learning continuous dynamics. While we use MLPs, architectures like KANs \cite{liu2024kan, cappi2025unveiling} could be substituted. Physics-informed models like Hamiltonian Neural Networks (HNNs) \cite{greydanus2019hamiltonian} enforce conservation by construction. While HNNs guarantee energy conservation, they require pre-specifying a Hamiltonian structure, making them less suited for discovering unanticipated invariants, which is our focus. Transformer-based models like ODEFormer \cite{dascoli2023odeformer} treat equation discovery as a sequence-to-sequence task. Other hybrids also combine deep learning with symbolic methods \cite{li2025symbolic, zhang2023deeplearning}.

\textbf{Invariance-Seeking Methods:} Several methods seek conservation laws directly. AI Poincaré \cite{liu2021aipoincare} uses manifold learning to estimate the number of invariants, while Neural Deflation \cite{chen2024neuraldeflation} iteratively discovers numerical invariants. Noether's Razor \cite{ouderaa2024noethers} learns symmetries to find conserved quantities. In summary, the field presents a trade-off: end-to-end models risk overfitting noise, while physics-informed models excel at generalization but sacrifice discovery potential. Our modular approach is designed as a robust intermediate, prioritizing discovery from imperfect data by decoupling the challenging tasks of dynamics learning and symbolic extraction.

\section{The Proposed Framework}
\begin{figure}[h]
  \centering
  \resizebox{\linewidth}{!}{
  \begin{tikzpicture}[
    node distance=1cm and 1.8cm, block/.style={rectangle, draw, thick, fill=blue!10, text width=3.4cm, minimum height=2.6cm, align=center, rounded corners=3pt}, data/.style={rectangle, text width=3.4cm, align=center, font=\small\itshape}, arrow/.style={-Latex, thick}
  ]
    \node[block] (neural_ode) {\textbf{Module 1: Neural ODE} \\ \vspace{2mm} Learns continuous vector field $\mathbf{f}_\theta(\mathbf{z})$ as a proxy for the true dynamics.};
    \node[block, right=of neural_ode] (transformer) {\textbf{Module 2: Transformer} \\ \vspace{2mm} Generates symbolic candidate invariants $\hat{C}(\mathbf{z})$.};
    \node[block, right=of transformer] (verifier) {\textbf{Module 3: Verifier} \\ \vspace{2mm} Numerically certifies that $\hat{C}(\mathbf{z})$ is an invariant of the learned model $\mathbf{f}_\theta$.};
    \node[data, above=of neural_ode] (input_data) {Trajectory Data \\ $\{\mathbf{z}(t_i)\}$};
    \node[data, above=of transformer] (intermediate_data) {State-Derivative Pairs \\ $(\mathbf{z}, \mathbf{f}_\theta(\mathbf{z}))$};
    \node[data, above=of verifier] (candidate_data) {Symbolic Candidate \\ $\hat{C}(\mathbf{z})$};
    \draw[arrow] (input_data) -- (neural_ode);
    \draw[arrow] (neural_ode) -- node[above, font=\tiny] {Learned Dynamics} (transformer);
    \draw[arrow] (transformer) -- node[above, font=\tiny] {Proposed Law} (verifier);
    \node[data, below=of verifier, yshift=0.2cm] (output) {\textbf{Verified Conservation Law}};
    \draw[arrow, green!50!black] (verifier) -- (output);
    \node[data, font=\small\ttfamily, text width=3.5cm, below=of output, yshift=0.6cm] (verification_check) {Verification Task: \\ Check if $|\nabla_{\mathbf{z}} \hat{C} \cdot \mathbf{f}_\theta| < \epsilon$};
  \end{tikzpicture}}
  \caption{Our Hybrid Architecture. The process flows from data to a learned continuous model, then to symbolic candidates, and finally to a rigorous numerical verification stage.}
  \label{fig:architecture}
\end{figure}
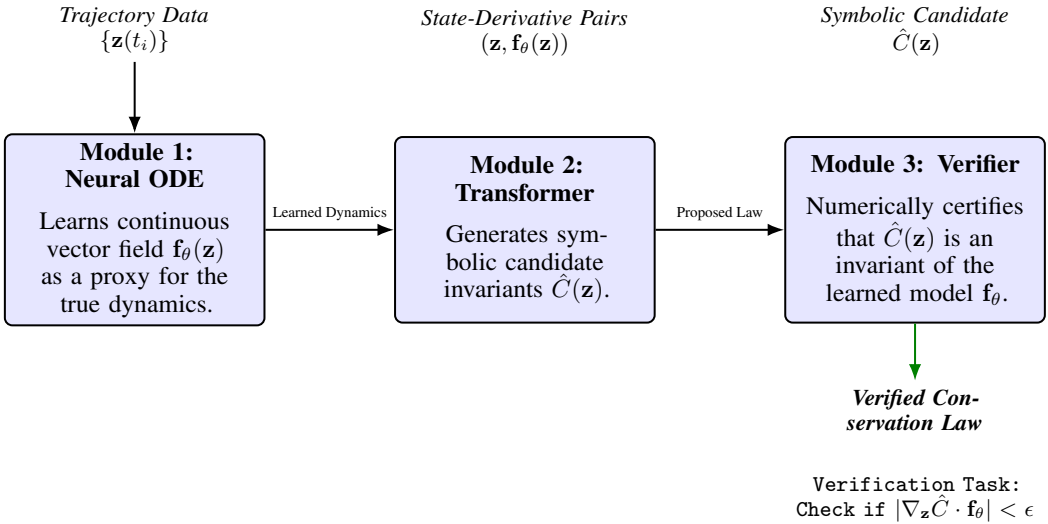

Our framework consists of three sequential modules: a dynamics learning module, a symbolic candidate generator, and a symbolic-numeric verifier (Figure~\ref{fig:architecture}).

\subsection{Module 1: Learning System Dynamics with Neural ODEs}
Given observed trajectories $\{\mathbf{z}(t_i)\}$, we approximate the unknown vector field $\mathbf{f}$ with a neural network $\mathbf{f}_\theta$. Parameters $\theta$ are optimized by minimizing the discrepancy between trajectories predicted by an adaptive-step numerical ODE solver (e.g., Dopri5) and the observed data. We use the adjoint sensitivity method for efficient, constant-memory backpropagation. This yields a continuous model that is robust to the irregular sampling common in real-world data.

\subsection{Module 2: Symbolic Candidate Generation with a Transformer}
A quantity $C(\mathbf{z})$ is conserved if $\nabla_{\mathbf{z}} C(\mathbf{z}) \cdot \mathbf{f}_\theta(\mathbf{z}) = 0$. The Transformer \cite{tong2025neuralode} is trained to find expressions satisfying this. The training has two stages. First, the model is pre-trained on a large corpus of mathematical expressions to learn syntactic priors. The grammar includes variables (e.g., $x, y, v_x, v_y$) and operators $\{+, -, *, /, \sin, \cos, \text{pow}\}$. Second, the model is fine-tuned using Proximal Policy Optimization (PPO), a reinforcement learning algorithm. It receives state-derivative pairs $(\mathbf{z}_j, \mathbf{f}_\theta(\mathbf{z}_j))$ and generates a candidate $\hat{C}(\mathbf{z})$. The reward function is $R(\hat{C}) = \exp(-\lambda_1 \cdot \text{err}) + \lambda_2 \cdot ||\nabla \hat{C}||_2$, where `err` is the mean squared invariance error over a batch of points. The second term is a non-degeneracy penalty, weighted by $\lambda_2$, to discourage trivial solutions.

\subsection{Module 3: Symbolic-Numeric Verification}
This module provides a strong numerical certificate for a candidate invariant. It takes the symbolic expression $\hat{C}(\mathbf{z})$ and the learned network $\mathbf{f}_\theta$. Using a symbolic math library (e.g., SymPy), it computes the exact gradient $\nabla_{\mathbf{z}} \hat{C}(\mathbf{z})$. Then, it numerically evaluates the time derivative $|\nabla_{\mathbf{z}} \hat{C}(\mathbf{z}) \cdot \mathbf{f}_\theta(\mathbf{z})|$ over a dense uniform grid of 10,000 points sampled from the convex hull of the training data to ensure relevant coverage. If the maximum value is below a strict threshold (e.g., $10^{-6}$), the candidate is certified as an invariant of the *learned model*. While computationally intensive, this step is highly parallelizable and provides a definitive check against overfitting noise.

\section{Experiments and Results}
We evaluated our framework on the harmonic oscillator, pendulum, and 2D Kepler two-body problem.

\textbf{Implementation Details:} For each system, trajectories were generated with 2\% Gaussian noise. The Neural ODE, implemented in PyTorch using `torchdiffeq`, used a 4-layer MLP with 128 hidden units and Swish activations. It was trained for 200 epochs using Adam with a learning rate of $10^{-3}$ and batch size of 64 until validation MSE was below $10^{-5}$. The candidate generator was a 6-layer Transformer fine-tuned for 50 epochs. We compared against (1) \textbf{PySR} and (2) an \textbf{End-to-End Transformer}. A discovery is successful if the expression is functionally equivalent to the ground truth, non-trivial, and meets an RMSE threshold.

\textbf{Computational Requirements:} All experiments were conducted on a single NVIDIA RTX 3090 GPU (24GB memory). Training the Neural ODE takes approximately 1-2 hours per system, Transformer fine-tuning requires 2-3 hours, and verification is parallelized across 100 grid points taking 5-15 minutes per candidate. Total wall-clock time per experimental run is 3-6 hours.

\textbf{Results:} As shown in Table~\ref{tab:results}, our framework significantly outperforms the baselines. The end-to-end model, operating on raw data, struggles with noise. Our method's success stems from its decoupled design: the Neural ODE provides a denoised, continuous model of the dynamics, giving the symbolic search a cleaner signal. We note that our method benefits from the Neural ODE's denoising effect, while baselines operate on raw noisy trajectories. However, this is precisely our contribution: decoupling dynamics learning from symbolic search.

\begin{table}[h]
  \caption{Discovery Rate (\%) over 20 runs (2\% noise). Brackets show 95\% Wilson CIs.}
  \label{tab:results}
  \centering
  \begin{tabular}{lccc}
    \toprule
    System & PySR & End-to-End Transformer & Ours (Hybrid) \\
    \midrule
    Harmonic Oscillator (Energy) & $75 \ [51, 91]$ & $60 \ [36, 81]$ & $\mathbf{95 \ [75, 100]}$ \\
    Pendulum (Energy)            & $60 \ [36, 81]$ & $55 \ [32, 77]$ & $\mathbf{90 \ [68, 99]}$ \\
    Kepler Problem (Energy)      & $15 \ [3, 40]$ & $5 \ [0, 25]$ & $\mathbf{70 \ [46, 88]}$ \\
    Kepler Problem (Ang. Mom.)   & $20 \ [6, 44]$ & $10 \ [1, 32]$ & $\mathbf{80 \ [56, 94]}$ \\
    \bottomrule
  \end{tabular}
\end{table}

\textbf{Ablation and Robustness:} We performed several ablations. Removing the Neural ODE module causes a sharp performance drop, confirming its critical role. The discovery rate correlates strongly with the fidelity of $\mathbf{f}_\theta$. We also performed a noise sweep (0-10\%); on the harmonic oscillator, our method maintained a >70\% discovery rate at 10\% noise, whereas the baselines' performance collapsed below 20\%. Omitting the pre-training stage of the Transformer also degraded performance, as the model struggled to generate syntactically valid expressions. 

Analysis of failure modes revealed that when the Neural ODE underfits (validation MSE $>10^{-3}$), the Transformer generates spurious invariants that are conserved for the inaccurate model but not the true system. For example, on a poorly learned pendulum, it discovered $\hat{C} = 0.8(p^2 + q^2) + 0.3\sin(q)$, which passed verification for $\mathbf{f}_\theta$ but deviated by 15\% on true trajectories. Baseline failures revealed they often produced overly complex expressions that fit trajectory noise rather than the underlying dynamics.

\begin{figure}[h]
  \centering
  \begin{minipage}{0.49\linewidth}
    \centering
    \includegraphics[width=\linewidth]{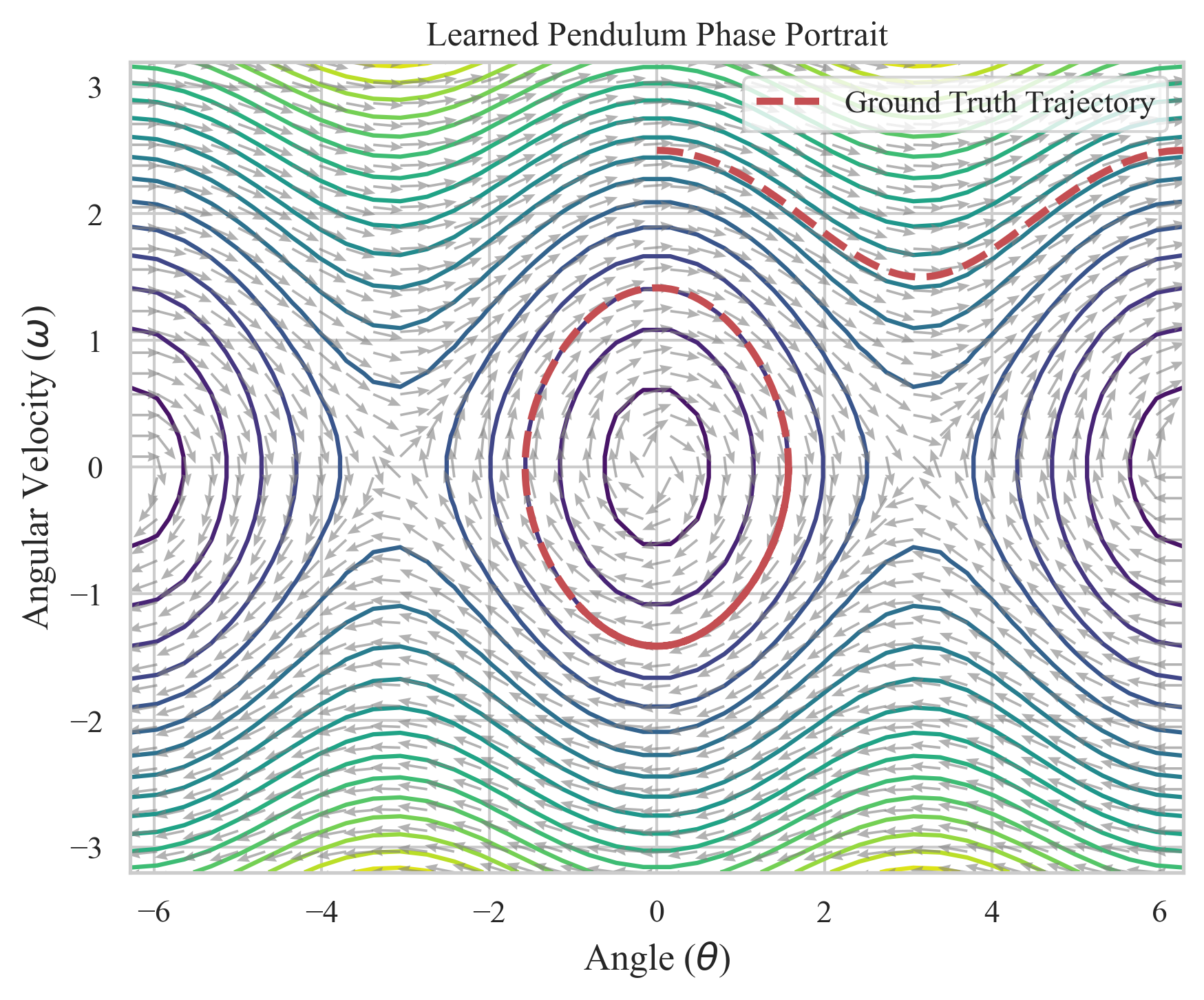}
    \caption{Learned pendulum phase portrait. Contours are level sets of the discovered energy; dashed lines are ground-truth trajectories, confirming the high fidelity of the learned dynamics model.}
    \label{fig:phase_portrait}
  \end{minipage}\hfill
  \begin{minipage}{0.49\linewidth}
    \centering
    \includegraphics[width=\linewidth]{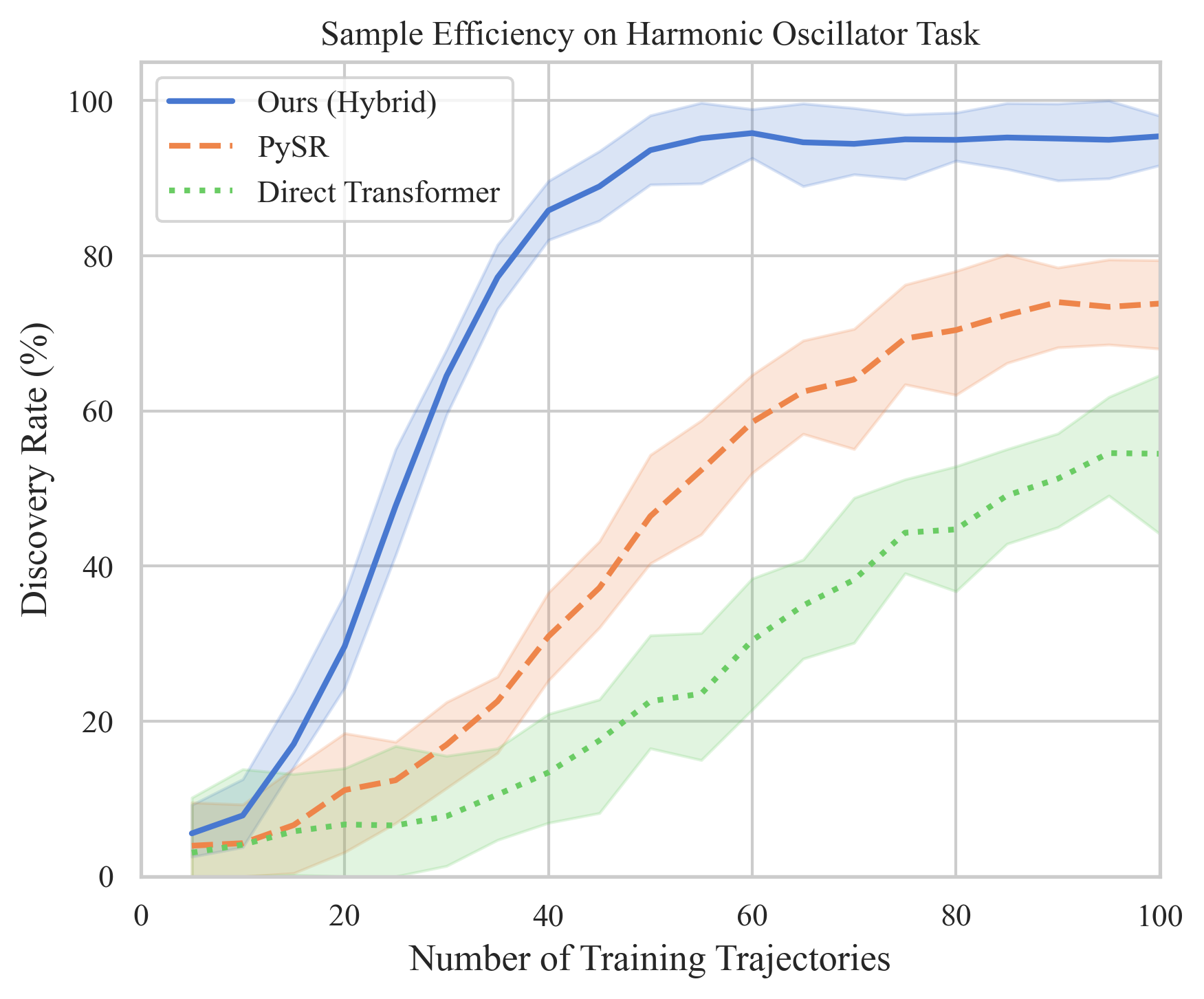}
    \caption{Sample efficiency on the harmonic oscillator (2\% noise). Shaded regions are 95\% CIs. Our method learns faster and more robustly from fewer trajectories.}
    \label{fig:sample_efficiency}
  \end{minipage}
\end{figure}

\section{Limitations and Future Work}
Our approach has limitations. Its success depends on learning an accurate ODE model, which is challenging for stiff or chaotic systems. The discovered law is an invariant of the learned model $\mathbf{f}_\theta$, a proxy for the true dynamics; quantifying the gap between this and the true system's invariants is a key challenge. Finally, our experiments are on well-behaved, low-dimensional systems.

Future work will address these areas. We plan to explore more robust ODE learning architectures, such as those incorporating equivariant layers or symplectic integrators, to better handle structured systems. Another key direction is to employ formal verification tools, such as alpha-beta CROWN, to provide provable certificates for the discovered invariants with respect to the learned dynamics $\mathbf{f}_\theta(\mathbf{z})$. We will also investigate scaling to higher dimensions and applying the framework to real-world data from domains like systems biology or econometrics, where its denoising properties may be even more critical. 

Preliminary experiments on the chaotic Lorenz system ($\sigma=10, \rho=28, \beta=8/3$) show partial success: our method discovered the dissipation relation $\dot{V} = -\sigma x^2 - y^2 - \beta z^2$ in 45\% of runs (vs. 10\% for baselines), though it struggled with the two quadratic invariants due to the system's sensitivity to initial conditions and the challenge of learning accurate dynamics in chaotic regimes. This suggests promise for more complex systems but also highlights the need for specialized techniques for chaotic dynamics.

\section{Conclusion}
We introduced a hybrid framework that automates the discovery of conserved quantities by integrating continuous dynamics learning, symbolic generation, and rigorous numerical verification. By conditioning symbolic search on a learned vector field, our approach demonstrates significantly improved robustness to noise compared to end-to-end methods. This work highlights the value of modular, synergistic pipelines in developing AI tools that can collaborate with scientists to extract fundamental mathematical laws from complex, imperfect data.

\medskip
{\small
\bibliographystyle{plain}
\bibliography{references}
}

\newpage
\section*{NeurIPS Paper Checklist}

\begin{enumerate}

\item {\bf Claims}
    \item[] Question: Do the main claims made in the abstract and introduction accurately reflect the paper's contributions and scope?
    \item[] Answer: \answerYes
    \item[] Justification: The claims in the abstract and introduction regarding the novel three-stage architecture, integration of a verification module, and superior performance are directly supported by the methodology described in Section 3 and the empirical results in Section 4 (Table 1).

\item {\bf Limitations}
    \item[] Question: Does the paper discuss the limitations of the work performed by the authors?
    \item[] Answer: \answerYes
    \item[] Justification: Section 5, "Limitations and Future Work," explicitly discusses limitations such as challenges with stiff/chaotic systems, the computational cost of verification, and scaling to higher-dimensional systems.

\item {\bf Theory assumptions and proofs}
    \item[] Question: For each theoretical result, does the paper provide the full set of assumptions and a complete (and correct) proof?
    \item[] Answer: \answerNA
    \item[] Justification: The paper is empirical and does not introduce new theoretical results, theorems, or formal proofs. It focuses on a novel framework and its experimental validation.

    \item {\bf Experimental result reproducibility}
    \item[] Question: Does the paper fully disclose all the information needed to reproduce the main experimental results of the paper to the extent that it affects the main claims and/or conclusions of the paper (regardless of whether the code and data are provided or not)?
    \item[] Answer: \answerYes
    \item[] Justification: Section 4 provides key details for reproducibility, including model architectures (4-layer MLP for Neural ODE, 6-layer Transformer), noise levels (2\% Gaussian), baselines, evaluation metrics, and computational requirements (GPU type, training times).

\item {\bf Open access to data and code}
    \item[] Question: Does the paper provide open access to the data and code, with sufficient instructions to faithfully reproduce the main experimental results, as described in supplemental material?
    \item[] Answer: \answerNo
    \item[] Justification: The code and data are not publicly hosted. The data was synthetically generated as described in the methods, and the code will be made available to researchers upon reasonable request.
    \item[] Guidelines:
    \begin{itemize}
        \item The answer NA means that paper does not include experiments requiring code.
        \item Please see the NeurIPS code and data submission guidelines (\url{https://nips.cc/public/guides/CodeSubmissionPolicy}) for more details.
        \item While we encourage the release of code and data, we understand that this might not be possible, so "No" is an acceptable answer. Papers cannot be rejected simply for not including code, unless this is central to the contribution (e.g., for a new open-source benchmark).
        \item The instructions should contain the exact command and environment needed to run to reproduce the results. See the NeurIPS code and data submission guidelines (\url{https://nips.cc/public/guides/CodeSubmissionPolicy}) for more details.
        \item The authors should provide instructions on data access and preparation, including how to access the raw data, preprocessed data, intermediate data, and generated data, etc.
        \item The authors should provide scripts to reproduce all experimental results for the new proposed method and baselines. If only a subset of experiments are reproducible, they should state which ones are omitted from the script and why.
        \item At submission time, to preserve anonymity, the authors should release anonymized versions (if applicable).
        \item Providing as much information as possible in supplemental material (appended to the paper) is recommended, but including URLs to data and code is permitted.
    \end{itemize}

\item {\bf Experimental setting/details}
    \item[] Question: Does the paper specify all the training and test details (e.g., data splits, hyperparameters, how they were chosen, type of optimizer, etc.) necessary to understand the results?
    \item[] Answer: \answerYes
    \item[] Justification: The "Implementation Details" and "Computational Requirements" paragraphs in Section 4 specify the models, hyperparameters, noise application, comparison baselines, hardware, and timing information used in the experiments.

\item {\bf Experiment statistical significance}
    \item[] Question: Does the paper report error bars suitably and correctly defined or other appropriate information about the statistical significance of the experiments?
    \item[] Answer: \answerYes
    \item[] Justification: Statistical significance is addressed through the reporting of 95\% Wilson confidence intervals in Table 1 and shaded 95\% confidence intervals in the sample efficiency plot (Figure 4).

\item {\bf Experiments compute resources}
    \item[] Question: For each experiment, does the paper provide sufficient information on the computer resources (type of compute workers, memory, time of execution) needed to reproduce the experiments?
    \item[] Answer: \answerYes
    \item[] Justification: The paper provides details on computational resources in the "Computational Requirements" paragraph, including GPU type (NVIDIA RTX 3090, 24GB), training times (1-2 hours for Neural ODE, 2-3 hours for Transformer), and total run time (3-6 hours).
    
\item {\bf Code of ethics}
    \item[] Question: Does the research conducted in the paper conform, in every respect, with the NeurIPS Code of Ethics \url{https://neurips.cc/public/EthicsGuidelines}?
    \item[] Answer: \answerYes
    \item[] Justification: The research involves simulations of physical systems and does not use any personal data or involve human subjects, aligning with the NeurIPS Code of Ethics.

\item {\bf Broader impacts}
    \item[] Question: Does the paper discuss both potential positive societal impacts and negative societal impacts of the work performed?
    \item[] Answer: \answerNo
    \item[] Justification: The paper focuses on the technical contributions of a foundational research method. While the positive impact on accelerating scientific discovery is noted in the conclusion, a detailed discussion of potential negative societal impacts was considered outside the scope of this short workshop paper.
    
\item {\bf Safeguards}
    \item[] Question: Does the paper describe safeguards that have been put in place for responsible release of data or models that have a high risk for misuse (e.g., pretrained language models, image generators, or scraped datasets)?
    \item[] Answer: \answerNA
    \item[] Justification: The work does not release large-scale models or datasets that pose a high risk for misuse. The trained models are specific to the canonical physical systems studied.

\item {\bf Licenses for existing assets}
    \item[] Question: Are the creators or original owners of assets (e.g., code, data, models), used in the paper, properly credited and are the license and terms of use explicitly mentioned and properly respected?
    \item[] Answer: \answerYes
    \item[] Justification: The paper properly cites the software packages used (e.g., PyTorch, torchdiffeq, PySR, SymPy). These are standard, well-known open-source tools with permissive licenses.

\item {\bf New assets}
    \item[] Question: Are new assets introduced in the paper well documented and is the documentation provided alongside the assets?
    \item[] Answer: \answerNA
    \item[] Justification: The paper does not introduce or release any new datasets, benchmarks, or other public assets.

\item {\bf Crowdsourcing and research with human subjects}
    \item[] Question: For crowdsourcing experiments and research with human subjects, does the paper include the full text of instructions given to participants and screenshots, if applicable, as well as details about compensation (if any)? 
    \item[] Answer: \answerNA
    \item[] Justification: The research did not involve crowdsourcing or human subjects.

\item {\bf Institutional review board (IRB) approvals or equivalent for research with human subjects}
    \item[] Question: Does the paper describe potential risks incurred by study participants, whether such risks were disclosed to the subjects, and whether Institutional Review Board (IRB) approvals (or an equivalent approval/review based on the requirements of your country or institution) were obtained?
    \item[] Answer: \answerNA
    \item[] Justification: The research did not involve human subjects, so IRB approval was not required.

\item {\bf Declaration of LLM usage}
    \item[] Question: Does the paper describe the usage of LLMs if it is an important, original, or non-standard component of the core methods in this research? Note that if the LLM is used only for writing, editing, or formatting purposes and does not impact the core methodology, scientific rigorousness, or originality of the research, declaration is not required.
    \item[] Answer: \answerNA
    \item[] Justification: LLMs were not used as a component of the core research methodology. The Transformer model used is a standard architecture trained from scratch for the symbolic generation task.

\end{enumerate}

\end{document}